# BAYESIAN INFERENCE FOR
# RADAR IMAGERY BASED SURVEILLANCE

Tod S. Levitt

Advanced Decision Systems
201 San Antonio Circle, Suite 286
Mountain View, California 94040

## 1. INTRODUCTION

We are interested in creating an automated or semi-automated system with the capability of taking a set of radar imagery, collection parameters and a priori map and other tactical data, and producing likely interpretations of the possible military situations given the available evidence. This paper is concerned with the problem of the interpretation and computation of certainty or belief in the conclusions reached by such a system. For example, if we consider the problem of confirming or denying the presence of a battalion in a given area, we should include in our decision making process the prior likelihood of military presence based on tactical objectives, the evidence of military vehicles in radar image data, the spatial and tactical clustering and patterns of the vehicles extracted from the imagery, etc. Furthermore, if the user of the system has particular interests such as knowing specific deployments, location of battalion headquarters, etc., then these interests should also be responded to in certainty computations.

In this report, we begin by briefly summarizing these functions from the point of view of outlining the inference processes that such a system must perform. Inference is performed over a space of hierarchically linked hypotheses. The hypotheses typically (although not solely) represent statements of the form "There is a military force of type $F$ in deployment $D$ at world location $L$ at time $T$". The hierarchy in the hypothesis space corresponds to the hierarchy inherent in military doctrine of force structuring. Thus, "array-level" hypotheses of military units such as companies, artillery batteries, and missile sites, are linked to their component unit hypotheses of vehicles batteries and missile launchers. Similarly, companies are grouped to form battalion hypotheses, battalions to form regiments, etc.

The hypotheses are formed by (hierarchical and partial) matching of military force models to evidence available in radar imagery. Thus, we are actually concerned with a model-based radar vision system. Evidence of the truth (or denial) of a hypothesis is accrued numerically from probabilistic estimates about the sub-hypotheses that comprise their parent hypothesis according to the hierarchical military force models in which the vehicle level hypotheses correspond to the leaf-nodes of the hierarchy. In this report, we only address the symbolic inference problem, under the assumption that probabilities of vehicle level hypotheses have already been computed. The vehicle level hypotheses are inferred from evidence supplied by radar image understanding algorithms applied to the radar data. These computations require radar modeling and development of a probabilistic certainty calculus, but these issues are not addressed here. See [Levitt et al., -86a] for an example of certainty calculus design at the vehicle level.

Summarizing so far, the proposed certainty calculus is a computational method for associating numerical probabilities to conclusions reached by a model-based radar vision system applied to problems of military situation assessment. A fundamental concept here is that, while vision system processing may be complex, with numerous feedback loops, multiple levels of resolution, recursion, etc., in the end we should be able to associate a deductive chain of evidence to a system output, along with an associated numerical belief that supports that result. The system concept is pictured in Figure 1.

We selected probability theory as the underlying technology for this numerical accrual of evidence. This approach requires us to lay out, a priori, the links between evidence and hypotheses in the models over which the system will reason. Having laid out these links, we then need a numerical interpretation of the conditional belief (i.e., probability) in a hypothesis given chains of evidence that support it through links. This is similar to the propagation networks of Pearl [Pearl - 85], to influence diagrams [Howard and Matheson - 80], and other probabilistic accrual models such as those of [Kelly and Barclay - 73], [Schum - 77, 80], and [Schum and Martin - 82]. Besides explicit declaration of evidence/hypothesis links, these approaches also require development of numerical accrual formula to capture the semantics of the links, and elicitation of a priori probabilities. However, a certainty calculus for perceptually based military situation assessment addresses additional issues not directly faced in these approaches.

In the aforementioned schemes, the network, influence diagram, etc. typically represents exactly one model at a single level of abstraction. (Pearl has extended his work to hierarchies [Pearl - 86], however, previous applications of his theory have not been hierarchical in nature. Schum's work is also hierarchical, but both are still concerned with a single model.) For example, a model might represent the chains of evidence from symptoms to disease for a diagnosis system. The model is "instantiated" for a single patient; the evidence flows through, and (numerical) conclusions are inferred. Conceptually we perform exactly one pattern match of a single model against a data set known to represent a unique instance (i.e., one person).

In this application there are multiple models linked hierarchically in multiple levels of abstraction. We have a basic (ascending) "part-of" hierarchy of military forces: "vehicle, part-of, company, part-of, battalion, part-of, regiment, part-of, division." Further, at each level of this hierarchy there is an "is-a" hierarchy of military force type refinements. For example, at the vehicle level we have: "T-72-tank, is-a, tank, is-a, tracked-vehicle, is-a, vehicle." Similarly, at the array level we have: tank-company-in-defensive deployment, is-a, tank-company, is-a, company, is-a, array." Thus, the model space represents multiple partial orderings at multiple resolution in both part-of and is-a hierarchies. (N.B. This is a well-known issue in knowledge representation in artificial intelligence. See, for example, [Brachman - 85].)

The second major difference is that we dynamically (i.e., at system runtime) generate multiple matches of models to instances in the radar image data. This corresponds to a hypothesis space with many hypotheses, more than one of which can be correct at the same time, e.g., there may be two divisions on the battlefield, each of which matches an instance of the same division model. This second difference is especially significant. The generation of multiple hypotheses at runtime gives rise to:

- the distribution of a finite amount of belief over a consistent set of hypotheses, and

- the problem of distinguishing conflicting interpretation of links of evidence to hypotheses (i.e., bad pattern matches) from consistent sets of multiple hypotheses.

The first issue is equivalent to not having a single "top level" model in the part-of hierarchy that bounds the number of parts (i.e., evidence) that can (consistently) occur below it. This problem can be finessed by deciding, a priori, that no more than a single division (for example) can be present in the data. In practice this sort of assumption is very reasonable, however, if we want to extend this approach to more general vision, to

168

adapting to changing spatial tactics by commanders, or to any sort of learning system, the requirement for a top level model must be removed.

The second issue is central in the current design. Conflict resolution between sets of hypotheses must be performed to attempt to improve the results of imperfect pattern matching. Matching models to the data is necessarily imperfect because, in general,

- only part of the forces actually present will be imaged in collected imagery.
- detection/recognition algorithms will miss some imaged vehicles, add in false alarms, and mis-classify some observed vehicles.
- models are not perfect and forces do not always follow precise doctrine, so incorrect matches will be made even when large parts of forces are observed.

Thus, the numerical accrual of evidence through the part-of and is-a hierarchies must not only account for accrual through consistent chains of inference, but also aid in disambiguating conflicting chains.

Multiple, possibly conflicting, model matches in a large data space give rise to combinatorial nightmares. We note, however, that the calculus only specifies how accrual should be performed over both conflicting and consistent sets of hypotheses. The control of the accrual process is technically a separable issue. Multiple control schemes could, in principle, be exercised over the same calculus, with radically different combinatorial results. This point is the springboard for the introduction of a technique for performing approximate hierarchical inference, while avoiding exponential processes. The technique presented is complementary to the (potentially expensive) conflict resolution procedures presented in [Levitt - 85] and [Levitt et al. - 86a].

Elicitation of highly conditional prior probabilities and likelihoods is another major research area necessary to design of the probability calculus. Priors combine domain expertise with background information beyond the knowledge boundaries (time, area, types of information) for which the automated system is responsible. We do not address this problem here. For a technical approach, see [Levitt et al. - 86b].

The structure of inference follows a pattern based on the models that are matched to generate hypotheses of the presence of military forces on the battlefield. These models consist purely of force types (i.e., names) and spatial-geometric deployment data relating force types. Modeling of the appearance of vehicles in SAR imagery, sensor models, terrain models, etc. are not addressed here. As depicted in Figure 1, each hypothesis passes on its posterior probabilities to become a priori (evidential) probabilities and hypotheses for hypotheses at the next level of hierarchical inference. Thus, certainties of vehicle detections are passed to inferences of vehicle classification, which in turn passes certainties of vehicle classifications to array-level military unit inference processes, etc. In general, posteriors at one force level become priors for their parent forces in the military hierarchy. Contextual analysis provides prior probabilities concerning a priori terrain evidence, likelihood of force presence based on a priori tactical considerations, as well as runtime estimates of accuracy of image-to-map registration that are necessary to account for the relative locations of forces which are observed in different sets of imagery.

In the following, we address the general problem of symbolic hierarchical Bayesian inference for military force inference. The key technical issue we address is the need to resolve conflicts between multiple incompatible hypotheses. The need to derive mutually consistent sets of (hierarchical) hypotheses gives rise to the exponential process of creating these sets. In this report, we present a method for approximate hierarchical accrual that can be used to selectively avoid unnecessary conflict resolution depending on the system's focus of attention in processing tasks.

169

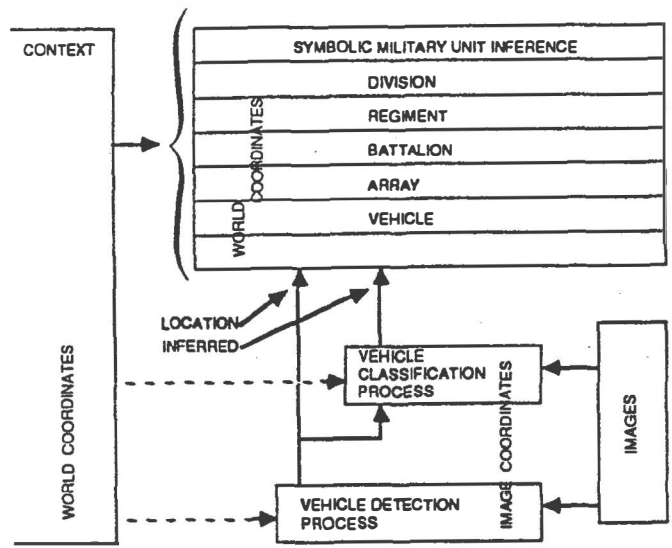

Figure 1: Hierarchical Inference Structure

## 2. EVIDENTIAL ACCRUAL

The vehicle classification process provides a hierarchy of probabilities concerning the hypothesis that a chip of synthetic aperture radar (SAR) imagery contains some vehicle from a well defined world, $W$, of vehicles. All probabilities are explicitly conditioned on contextual information including terrain, and radar dependent measurements, $t,m$, provided with the chip, and upon evidence, $e$, extracted from the chip. The vehicle classification process apportions the probability that a vehicle is present among the nodes of its is-a hierarchy. This process ends at the leaf nodes of the tree that represent specific vehicle types. The fundamental problem is to estimate the probability that $V$ occurs given evidence $e$ and context $m$ and $t$; that is, to compute $P(V \mid e,t,m)$. This is a deep radar modeling issue and we do not address it in this report. In the following we assume this computation has been performed.

We now address evidence accrual up the hierarchy of military force hypotheses. For each hypothesis, $H$, which is based on a model that has components, $C_i$, we first find the components which match the component portions of the model and their corresponding evidence, $e_i$, that supports each component hypothesis $C_i$. There is also a set of spatial pattern matching evidence associated with $H$ that is specific to the geometric, spatial constraints that are given by the model for $H$. This set of evidence is denoted $f$. A hypothesis with components is then defined from the model as the existence of a force that is composed of the listed components $C_i$. The evidence that supports the hypothesis $H$ is defined as the and of the evidence, $e_i$, for all of the component hypotheses, $C_i$, and the new evidence, $f$, for the hypothesis $H$. Initially, the $e_i$ are the image data and processing results and $H$ corresponds to a vehicle hypothesis based on that evidence. At the next level of the part-of hierarchy, the array level matches of vehicle sets give rise to array level hypotheses, $H$, the vehicle hypotheses play the role of components of $H$ in the part-of hierarchy, i.e., the $C_i$, and the image evidence represents the $e_i$. With this recursive definition the new set of evidence for the hypothesis $H$ will become the evidence that supports this hypothesis as it is used as a component of a yet higher level hypothesis that matches a yet higher level force model. We use as notation for the conjunction of the component hypotheses, $\bigwedge_i C_i$. This represents the statement that all of the component forces, $C_i$, exist. This does not necessarily imply that the hypothesis $H$ exists.

170

In numerical accrual of evidence, we make several independence assumptions. The first is the standard hierarchical inference assumption that deals with evidence about an intermediate level hypothesis, $C$, where there is a higher level hypothesis, $H$, that uses $C$ on the part-of hierarchy. That is, there exists evidence that infers $C$ and then $C$ infers $H$. The assumption is that given that $C$ is true, the evidence, $e$ supporting $C$ and $H$ are conditionally independent. That is $P(H, e \mid C) = P(H \mid C) \cdot P(e \mid C)$.

The second assumption concerns the independence of evidence at various levels in a consistent hierarchy. If there is a hypothesis that is supported both by components and by inter-relations between those components, and if we assume that the hypothesis is true, then the evidence of the relations among the components is independent of the evidence for each component. This can be written as a single statement as follows. Let the $e_i$'s be the evidence supporting each component for a hypothesis $H$; then

$$P(\wedge_i e_i \mid H, \wedge_i C_i) = \prod_i P(e_i \mid H, C_i)$$

Here "$\wedge_i e_i$" means the conjunction of the evidence supporting the components.

We also assume that if $C$ is a part-of the hypothesis $H$, then $P(C \mid H) = 1$.

The next independence assumption concerns conflict recognition between hypotheses. If $C_1$, $C_2$ are hypotheses supported by disjoint sets of evidence $e_1$ and $e_2$ respectively and, furthermore, $C_1$ and $C_2$ are not in military doctrinal conflict (e.g., by being situated too close together, facing opposite directions, etc.), then:

$$P(C_1 \mid e_1, e_2, C_2) = P(C_1 \mid e_1).$$

Now, if $\wedge_i C_i$ are parts-of $H$, and the $\wedge_i C_i$ are supported by non-conflicting evidence $\wedge_i e_i$, then the value we need to accrue is $P(H \mid \wedge_i C_i, \wedge_i e_i, f, \wedge_i t_i)$. Where $t_i$ is the terrain under $C_i$, and $f$ is a measure of the fit to formation of the $C_i$ to the parent hypothesis $H$. In doing so, we also assume that $P(t_i) = 1$, that is, we know the world terrain with absolute certainty, and also that terrain is a local issue so that $P(\wedge_i t_i \mid H, \wedge_i C_i, \wedge_i e_i, f) = \prod_i (t_i \mid C_i)$. Using Bayes rule under these assumptions, we can show that:

$$P(H \mid \wedge_i C_i, \wedge_i e_i, t_i, f) = \frac{P(f \mid H, \wedge_i C_i)}{P(f \mid \wedge_i e_i, \wedge_i t_i)} \cdot \prod_i \left[ \frac{P(C_i \mid e_i) P(C_i \mid t_i)}{P(C_i \mid e_i, t_i)} \cdot \frac{P(H)}{P^2(C_i)} \right]$$

## 3. APPROXIMATE CONFLICT RESOLUTION

The assumption in the previous section that $P(C_1 \mid e_1, e_2, C_2) = P(C_1 \mid e_1)$ is clearly unacceptable if either $e_1$ and $e_2$ are not disjoint, or if the truth of $C_2$ conflicts for any reason with that of $C_1$. A methodology for handling conflicts was presented in [Levitt - 85]. The concept was to accrue non-conflicted evidence up to the highest level of the hierarchy, and then to form a mutually exclusive, exhaustive hypothesis space of maximal consistent sets of hypotheses. Results of disambiguation at the highest level could be passed down the hierarchy with high probability of correct inference. An alternative approach is presented in [Levitt et al. - 86a] where conflict resolution is performed bottom-up in the accrual process by factoring in the negation of each hypothesis in the accrual. Both approaches suffer from an inherently worst-case exponential step in forming sets of consistent hypotheses.

Here, we present a method for analyzing a conflict in polynomial time, and provide a metric, based on the analysis, to determine if conflict is worth dis-ambiguating. If it is not, then accrual must be able to "jump" over the conflicted level in the hierarchy by specifying how to accrue evidence at lower levels as support for higher level hypotheses.

The concept of conflict analysis is embodied in the case for two hypotheses. Let $e = e_1 \cup e_2 \cup e_{12}$ where $e_1$ is a set of evidence supporting only $C_1$, $e_2$ supports only $C_2$ and $e_{12}$ is shared, conflicted evidence associated to both. Then we approximate

$$P(C_1, C_2, e) = P(C_1 \mid C_2, e) P(C_2 \mid e) P(e)$$
$$\approx P(C_1 \mid e_1) P(C_2 \mid e_2 \cup e_{12}) P(e)$$
$$\approx P(C_2 \mid e_2) P(C_1 \mid e_1 \cup e_{12}) P(e).$$

The underlying independence assumptions are that $P(C_1 \mid e_2) = P(C_1)$ and $P(C_2 \mid e_1) = P(C_2)$. During control of inference, we use this in the following way. Let $\{C_i\}$ be a set of conflicted hypotheses, and let $e = \cup e_i$ be the total set of supporting evidence for the $C_i$. Use heuristics (e.g., most matches, highest total a priori probability, etc.) to order the $C_i$. Then form the evidence sets $\{e - \bigcup_{k=i}^{n} e_k\}$ for $i = 2$ to $n$. This is a polynomial time operation. Then approximate

$$P(\Lambda_i C_i \mid e) \approx \prod_{i=1}^{n} P(C_i \mid e - \bigcup_{k=i+1}^{n} e_i) = k.$$

$k$ is interpreted as a measure of how likely the $\{C_i\}$ are all to be true, despite the conflict in evidence association. We use $\dfrac{1-k}{k}$ as the measure of conflict. If it is small, then the error in not dis-ambiguating the conflict (i.e., accrual skipping the conflicted level in the hierarchy) will be small.

To see this, let $\Lambda_i C_i$ be associated to $H$, supported by $e$, in which, for convenience, we have included $\Lambda_i e_i$, $f$ and $\Lambda_i t_i$. Then $P(H \mid \Lambda_i C_i, e) = P(e \mid H) P(H) / [P(\Lambda_i C_i \mid e) P(e)]$ because $P(\Lambda_i C_i \mid H, e) = 1$. We have the error in "skipping" the level of the $C_i$ as

$$\mid P(H \mid \Lambda_i C_i, e) - P(H \mid e) \mid$$

$$= \mid \frac{P(e \mid H) P(H)}{P(\Lambda_i C_i \mid e) P(e)} - \frac{P(e \mid H) P(H)}{P(e)} \mid$$

$$= \frac{P(e \mid H) P(H)}{P(e)} \mid \frac{1}{P(\Lambda_i C_i \mid e)} - 1 \mid$$

$$= \frac{P(e \mid H) P(H)}{P(e)} \mid \frac{1 - P(\Lambda_i C_i \mid e)}{P(\Lambda_i C_i \mid e)} \mid$$

172

$$\approx \frac{P(e \mid H) P(H)}{P(e)} \left( \frac{1 - k}{k} \right)$$

which is small if $\frac{1-k}{k}$ is small.

## 4. ISSUES

Approximate hierarchical accrual provides a tool for choosing to ignore local conflicts in hierarchical inference. There are considerable issues in how to utilize it in the framework of system control. In this report we have presented a partial design for a probabilistic certainty calculus to support inference for military force inference. In the course of performing such inference, the step of exact conflict resolution will often be necessary. The inference control issue is how to decide when, in the face of conflict, to test for the use of approximate accrual and how to threshold the outcome of the test.

The current design of a certainty calculus requires considerable effort in refinement and implementation. Large grain work chunks include:

- Development of a control methodology to address computational combinatorics in the hierarchical inference evidence propagation and (approximate) conflict resolution.
- Elicitation and verification of distributions characterizing the relative occurrence of military forces versus terrain and military contextual knowledge.

## 5. ACKNOWLEDGEMENTS

I am indebted to Ward Edwards, Bob Kirby, and Larry Winter for extended discussions that helped clarify many of the points in this paper. Dubious semantic interpretations or errors are, of course, the author's sole responsibility.

This work was supported by Defense Advanced Research Projects Agency and U.S. Army Engineering Topographic Laboratories under U.S. Government Contract No. DACA76-86-C-0010, ADS Contract No. 1131.